\PassOptionsToPackage{table}{xcolor}
\documentclass[runningheads]{llncs}

\usepackage{eccv}
\usepackage{eccvabbrv}

\usepackage{graphicx}
\usepackage{booktabs}
\usepackage{amsmath}

\definecolor{cvprblue}{rgb}{0.21,0.49,0.74}
\definecolor{lightgray}{rgb}{0.83, 0.83, 0.83}
\definecolor{mossgreen2}{rgb}{0.58, 0.80, 0.48}
\definecolor{chromeyellow}{rgb}{1.0, 0.65, 0.0}
\definecolor{babyblue}{rgb}{0.54, 0.81, 0.94}
\definecolor{dark}{gray}{0.2}
\usepackage[breaklinks,colorlinks,citecolor=cvprblue]{hyperref}
\usepackage{pgfplots}
\usepackage{pgfplotstable}
\pgfplotsset{compat=1.8}

\usepackage{multirow}
\usepackage{makecell}

\newlength\savewidth\newcommand\shline{\noalign{\global\savewidth\arrayrulewidth
  \global\arrayrulewidth 1pt}\hline\noalign{\global\arrayrulewidth\savewidth}}

\usepackage[accsupp]{axessibility}

\usepackage{orcidlink}

\begin{document}

\title{LW-DETR: A Transformer Replacement to YOLO for Real-Time Detection}

\titlerunning{LW-DETR}

\author{Qiang Chen\inst{1}\thanks{Equal contribution. \quad$^{\dag}$Corresponding author.} \and
Xiangbo Su\inst{1*} \and
Xinyu Zhang\inst{2,1*} \and
Jian Wang\inst{1} \and
Jiahui Chen\inst{3,1} \and
Yunpeng Shen\inst{1} \and
Chuchu Han\inst{1} \and
Ziliang Chen\inst{1} \and
Weixiang Xu\inst{1} \and
Fanrong Li\inst{4} \and
Shan Zhang\inst{5} \and
Kun Yao\inst{1} \and
Errui Ding\inst{1} \and
Gang Zhang\inst{1} \and
Jingdong Wang\inst{1\dag}
}

\authorrunning{Q.~Chen et al.}

\institute{Baidu Inc., China \and
The University of Adelaide, Australia \and
Beihang University, China \and
Institute of Automation, Chinese Academy of Sciences, China \and
Australian National University, Australia \\
\email{\{chenqiang13,suxiangbo,zhangxinyu14,wangjingdong\}@baidu.com}}

\maketitle

\begin{abstract}
  In this paper, we present a light-weight detection transformer, LW-DETR, which outperforms YOLOs for real-time object detection. The architecture is a simple stack of a ViT encoder, a projector, and a shallow DETR decoder. Our approach leverages recent advanced techniques, such as training-effective techniques, e.g., improved loss and pretraining, and interleaved window and global attentions for reducing the ViT encoder complexity. We improve the ViT encoder by aggregating multi-level feature maps, and the intermediate and final feature maps in the ViT encoder, forming richer feature maps, and introduce window-major feature map organization for improving the efficiency of interleaved attention computation. Experimental results demonstrate that the proposed approach is superior over existing real-time detectors, e.g., YOLO and its variants, on COCO and other benchmark datasets. Code and models are available at \url{https://github.com/Atten4Vis/LW-DETR}.
  \keywords{Object Detection \and Real-Time \and Detection Transformer}
\end{abstract}

\begin{figure}[t]
\centering
\begin{tikzpicture}[font=\footnotesize]
\begin{axis}[
legend columns=1, 
legend style={at={(0.8,0.6)},anchor=north, font=\tiny, draw=gray!20}, legend cell align={left},
y label style={at={(-0.1,0.5)}},
ylabel={mAP}, xlabel={Inference Time (ms)}, ymajorgrids=true,
extra y ticks={59.0},
tick style={draw=none},
y label style={font=\footnotesize},
height=8cm,
width=9.5cm,
axis lines = left,
every outer y axis line/.style={draw=gray!40},
every outer x axis line/.style={draw=gray!40},
grid style={line width=.1pt, draw=gray!20},
xmin=0, ymax=59.0, ymin=36.5]

\addplot[line width=1.0pt, mark size=1.5pt, mark=diamond, draw=red, draw opacity=0.8]
table
{
X Y
2.0 42.6
2.9 48.0
5.6 52.5
8.8 56.1
19.1 58.3
};

\addplot[line width=1.0pt, mark size=1.5pt, mark=x, draw=black!80!white, draw opacity=0.8]
table
{
X Y
2.9 47.3
5.7 51.1
7.6 51.9
};

\addplot[line width=1.0pt, mark size=1.5pt, mark=x, draw=blue!80!white, draw opacity=0.8]
table
{
X Y
1.6 37.5
2.7 45.1
6.0 50.4
9.4 53.0
15.0 54.1
};

\addplot[line width=1.0pt, mark size=1.5pt, mark=x, draw=green!80!white, draw opacity=0.8]
table
{
X Y
2.4 41.5
2.9 44.7
6.5 49.5
10.5 52.2
18.8 53.5
};

\addplot[line width=1.0pt, mark size=1.5pt, mark=x, draw=black!30!white, draw opacity=0.8]
table
{
X Y
4.7 47.6
7.8 51.6
8.8 52.3
};

\addplot[line width=1.0pt, mark size=1.5pt, mark=x, draw=blue!30!white, draw opacity=0.5]
table
{
X Y
6.2 37.6
7.0 45.2
10.1 50.6
13.2 53.3
19.1 54.5
};

\addplot[line width=1.0pt, mark size=1.5pt, mark=x, draw=green!30!white, draw opacity=0.8]
table
{
X Y
7.4 41.7
7.9 44.9
10.8 49.7
14.9 52.4
22.8 54.0
};

\addlegendentry{LW-DETR} 
\addlegendentry{YOLO-NAS*} 
\addlegendentry{YOLOv8*} 
\addlegendentry{RTMDet*}  
\addlegendentry{YOLO-NAS} 
\addlegendentry{YOLOv8} 
\addlegendentry{RTMDet}
\end{axis}
\end{tikzpicture}
\caption{\textbf{Our approach outperforms previous SoTA real-time detectors.} The x-axis corresponds to the inference time. The y-axis corresponds to the mAP score on COCO \texttt{val2017}. 
All the models are trained with pretraining on Objects365.
The NMS post-processing times are included for other models and measured on the COCO \texttt{val2017} with the setting from the official implementation~\cite{lyu2022rtmdet,yolov8_ultralytics,supergradients}, and the well-tuned NMS postprocessing setting (labeled as ``*'').}
\label{fig:latencyvsperformance}
\vspace{-4mm}
\end{figure}
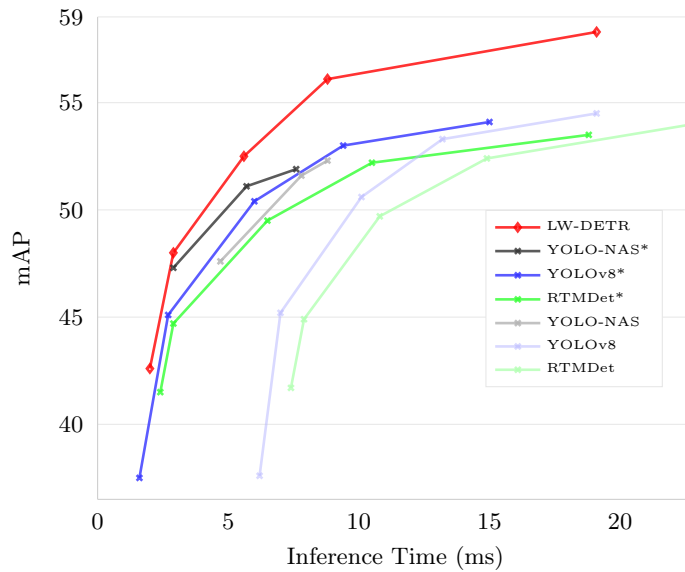

\section{Introduction}
\label{sec:intro}

Real-time object detection is an important problem in visual recognition and has wide real-world applications. The current dominant solutions are based on convolutional networks, such as the YOLO series~\cite{bochkovskiy2020yolov4,ge2021yolox,xu2022pp,wang2023yolov7,li2022yolov6,li2023yolov6,lyu2022rtmdet,yolov8_ultralytics,supergradients}. Recently, transformer methods, e.g., detection transformer (DETR)~\cite{carion2020end}, have witnessed significant progress~\cite{zhu2020deformable,meng2021conditional,gao2022adamixer,wang2022anchor,liu2022dab,zhang2022dino,chen2022group,zong2023detrs}. Unfortunately, DETR for real-time detection remains not fully explored, and it is unclear if the performance is comparable to the state-of-the-art convolutional methods.

In this paper, we build a light-weight DETR approach for real-time object detection. The architecture is very simple: a plain ViT encoder~\cite{dosovitskiy2020image} and a DETR decoder that are connected by a convolutional projector~\cite{yolov8_ultralytics}. We propose to aggregate the multi-level feature maps, the intermediate and final feature maps in the encoder, forming stronger encoded feature maps. Our approach takes advantage of effective training techniques. For example, we use the deformable cross-attention forming the decoder~\cite{zhu2020deformable}, the IoU-aware classification loss~\cite{cai2023align}, and the encoder-decoder pretraining strategy~\cite{zhang2023cae,zhang2022dino,chen2022group}.

On the other hand, our approach exploits inference-efficient techniques. For example, we adopt interleaved window and global attentions~\cite{li2021benchmarking,li2022exploring}, replacing some global attentions with window attentions in the plain ViT encoder to reduce the complexity. We use an efficient implementation for the interleaved attentions through a window-major feature map organization method, effectively reducing the costly memory permutation operations.

Figure~\ref{fig:latencyvsperformance} shows that the proposed simple baseline surprisingly outperforms the previous real-time detectors on COCO~\cite{lin2014microsoft}, \emph{e.g.}, YOLO-NAS~\cite{supergradients}, YOLOv8~\cite{yolov8_ultralytics}, and RTMDet~\cite{lyu2022rtmdet}. These models are improved with pretraining on Objects365~\cite{shao2019objects365}, and the end-to-end time cost, including the NMS time, is measured using the setting from the official implementations\footnote{\tiny
YOLO-NAS: \url{https://github.com/Deci-AI/super-gradients}\\
YOLOv8: \url{https://docs.ultralytics.com}\\
RTMDet: \url{https://github.com/open-mmlab/mmyolo/tree/main/configs/rtmdet}
}.

We conduct extensive experiments for the comparisons with existing real-time detection algorithms~\cite{supergradients,yolov8_ultralytics,lyu2022rtmdet,chen2023yolo,wang2023gold}. We further optimize the NMS setting and obtain improved performance for existing algorithms. The proposed baseline still outperforms these algorithms (labeled as ``*'' in Figure~\ref{fig:latencyvsperformance}). In addition, we demonstrate the proposed approach with experimental results on more detection benchmarks.

The proposed baseline merely explores simple and easily implemented techniques and shows promising performance. We believe that our approach potentially benefits from other designs, such as efficient multi-scale feature fusion~\cite{li2023lite}, token sparsification~\cite{zheng2023less,li2022exploring}, distillation~\cite{chang2023detrdistill,chen2022d,chang2023detrdistill,wang2022knowledge}, as well as other training techniques, such as the techniques used in YOLO-NAS~\cite{supergradients}. We also show that the proposed approach is applicable to the DETR approach with the convolutional encoder, such as ResNet-18 and ResNet-50~\cite{he2016deep}, and achieves good performance.

\section{Related Work}
\noindent\textbf{Real-time object detection.} 
Real-time object detection has wide real-world applications~\cite{li2019gs3d,feng2020deep,hu2023planning,karaoguz2019object,paul2021object}. 
Existing state-of-the-art real-time detectors, such as YOLO-NAS~\cite{supergradients}, YOLOv8~\cite{yolov8_ultralytics}, and RTMDet~\cite{lyu2022rtmdet}, have been largely improved compared with the first version of YOLO~\cite{redmon2016you} through
detection frameworks~\cite{tian2019fcos,ge2021yolox}, architecture designs~\cite{bochkovskiy2020yolov4,li2022yolov6,li2023yolov6,xu2022pp,ding2021repvgg,wang2023yolov7}, data augmentations~\cite{zhang2017mixup,bochkovskiy2020yolov4,ge2021yolox}, training techniques~\cite{ge2021yolox,yolov8_ultralytics,supergradients}, and loss functions~\cite{lin2017focal,zhang2021varifocalnet,zheng2020distance}. 
These detectors are based 
on convolutions.
In this paper,
we study transformer-based solutions to real-time detection that remains little explored.
\vspace{1mm}

\noindent\textbf{ViT for object detection.} 
Vision Transformer (ViT)~\cite{dosovitskiy2020image,zhai2022scaling} shows promising performance in image classification.
Applying ViT to object detection 
usually exploits
window attentions~\cite{dosovitskiy2020image,liu2021swin}
or hierarchical architectures~\cite{wang2021pyramid,liu2021swin,zhang2022hivit,hatamizadeh2023fastervit}
to reduce the memory
and computation cost.
UViT~\cite{chen2021simple} 
uses progressive window attention.
ViTDet~\cite{li2022exploring} implements the pre-trained plain ViT with interleaved window and global attentions~\cite{li2021benchmarking}. 
Our approach follows ViTDet 
to use interleaved window
and global attentions,
and additionally uses window-major order feature map organization
for reducing the memory permutation cost. 

\vspace{1mm}
\noindent\textbf{DETR and its variants.} 
Detection Transformer (DETR)
is an end-to-end
detection method,
with removing the necessity
of many hand-crafted components, such as anchor generation~\cite{ren2015faster} and non-maximum suppression (NMS)~\cite{hosang2017learning}. 
There are many followup methods
for DETR improvement,
such as 
architecture
design~\cite{zhu2020deformable,meng2021conditional,gao2022adamixer,zhang2022dino}, 
object query design~\cite{wang2022anchor,liu2022dab,chen2022conditional}, training techniques~\cite{li2022dn,zhang2022dino,chen2023group,jia2023detrs,ouyang2022nms,chen2022group,zong2023detrs}, 
and loss function improvement~\cite{liu2023detection,cai2023align}. 
Besides,
various works have been done
for reducing the computational complexity, 
by architecture design~\cite{lin2023detr,li2023lite}, computational optimization~\cite{zhu2020deformable}, pruning~\cite{roh2021sparse,zheng2023less}, and distillation~\cite{chang2023detrdistill,chen2022d,chang2023detrdistill,wang2022knowledge}. 
The interest 
of this paper
is 
to build a simple DETR baseline
for 
real-time detection 
that are not explored by these methods.

Concurrent with our work, RT-DETR~\cite{lv2023detrs} also applies the DETR framework to construct real-time detectors with a focus on
the CNN backbone forming the encoder.
There are studies about relatively large models
and a lack of tiny models. 
Our LW-DETR explores the feasibility of plain ViT backbones and DETR framework for real-time detection .

\begin{figure}[t]
\centering
\includegraphics[scale=1.0]{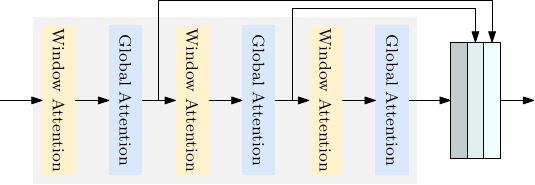}
\caption{\textbf{An example of transformer encoder} with 
multi-level feature map aggregation
and interleaved window and global attentions. The FFN and LayerNorm layers are not depicted for clarification.}
\label{fig:encoder}
\vspace{-3mm}
\end{figure}

\section{LW-DETR} \label{sec:LW-DETR}
\subsection{Architecture}
\label{sec:LW-DETR:Architecture}
LW-DETR consists of a ViT encoder, a projector, and a DETR decoder.

\vspace{1mm}
\noindent\textbf{Encoder.} We adopt the ViT 
for the detection encoder. A plain ViT~\cite{dosovitskiy2020image} consists of a patchification layer and transformer encoder layers. 
A transformer encoder layer in the initial ViT contains a global self-attention layer over all the tokens and an FFN layer. 
The global self-attention is computationally costly, and its time complexity is quadratic with respect to the number of tokens (patches). We implement some transformer encoder layers with window self-attention to reduce the computational complexity (detailed in Sec.~\ref{sec:efficient_inference}). 
We propose to aggregate the multi-level feature maps, the intermediate and final feature maps in the encoder, forming stronger encoded feature maps. An example of the encoder is illustrated in Figure~\ref{fig:encoder}.

\vspace{1mm}
\noindent\textbf{Decoder.} 
The decoder is a stack of transformer decoder layers. Each layer consists of a self-attention, a cross-attention, and an FFN. 
We adopt deformable cross-attention~\cite{zhu2020deformable} for computational efficiency. 
DETR and its variants usually adopt $6$ decoder layers. In our implementation, we use $3$ transformer decoder layers. 
This leads to a time reduction from $1.4$ ms to $0.7$ ms, 
which is significant compared to the time cost $1.3$ ms of the remaining part for the tiny version in our approach.

We adopt a mixed-query selection scheme~\cite{zhang2022dino} to form the object queries as an addition of content queries and spatial queries. The content queries are learnable embeddings, which is similar to DETR. 
The spatial queries are based 
on a two-stage scheme: selecting top-$K$ features 
from the last layer in the Projector, predicting the bounding boxes,  
and transforming the corresponding boxes into embeddings as spatial queries.

\vspace{1mm}
\noindent\textbf{Projector.} We use a projector to connect the encoder and the decoder. The projector takes the aggregated encoded feature maps from the encoder as the input. The projector is a C2f block (an extension of cross-stage partial DenseNet~\cite{huang2017densely,wang2020cspnet}) that is implemented in YOLOv8~\cite{yolov8_ultralytics}.

\begin{figure}[t]
\centering
\footnotesize
\includegraphics[scale=1.0]{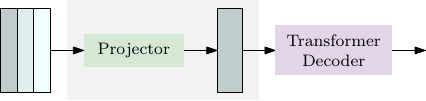}\\
(a)\\
\vspace{2mm}
\includegraphics[scale=1.0]{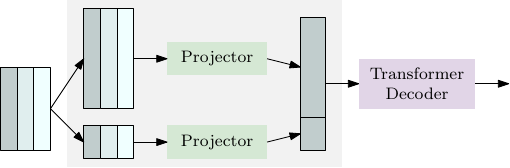}\\
(b)\\
\caption{\textbf{Single-scale projector and multi-scale projector} for (a) the \texttt{tiny}, \texttt{small}, and \texttt{medium} models, and (b) the \texttt{large} and \texttt{xlarge} models.}
\label{fig:multicaleprojector}
\vspace{-3mm}
\end{figure}

When forming the \texttt{large} and \texttt{xlarge} version of LW-DETR, we modify the projector to output two-scale ($\frac{1}{8}$ and $\frac{1}{32}$) feature maps and accordingly use the multi-scale decoder~\cite{zhu2020deformable}. The projector contains two parallel C2f blocks. One processes $\frac{1}{8}$ feature maps, which are obtained by upsampling the input through a deconvolution, and the other processes $\frac{1}{32}$ maps that are obtained by downsampling the input through a stride convolution. Figure~\ref{fig:multicaleprojector} shows the pipelines of the single-scale projector and the multi-scale projector.

\vspace{1mm}
\noindent\textbf{Objective function.} We adopt an {IoU-aware classification} loss, IA-BCE loss~\cite{cai2023align},
\begin{align}
\ell_{\texttt{cls}} = \sum_{i=1}^{N_{pos}}\operatorname{BCE}(s_i, t_i) + \sum_{j=1}^{N_{neg}} s_j^2\operatorname{BCE}(s_j,0),
\end{align}
where $N_{pos}$ and $N_{neg}$ are the number of positive and negative samples. $s$ is the predicted classification score. $t$ is the target score absorbing the IoU score $u$ (with the ground truth): $t=s^{\alpha} u^{1-\alpha}$, and $\alpha$ is empirically set as $0.25$~\cite{cai2023align}. 

The overall loss is a combination of the classification loss and the bounding box loss that is the same as in the DETR frameworks~\cite{carion2020end,zhu2020deformable,zhang2022dino}, which is formulated as follows:
\begin{align}
    \ell_{\texttt{cls}} + \lambda_{\texttt{iou}}\ell_{\texttt{iou}} + \lambda_{\ell_1}\ell_1.
\end{align}
where $\lambda_{\texttt{iou}}$ and $\lambda_{\ell_1}$ are set as $2.0$ and $5.0$ similar to~\cite{carion2020end,zhu2020deformable,zhang2022dino}. $\ell_{\texttt{iou}}$ and $\ell_1$ are the generalized IoU (GIoU) loss~\cite{rezatofighi2019generalized} and the L$1$ loss for the box regression.

\subsection{Instantiation}
We instantiate $5$ real-time detectors: \texttt{tiny}, \texttt{small}, \texttt{medium}, \texttt{large}, and \texttt{xlarge}. The detailed settings are given in Table~\ref{tab:modelscaling}.

The \texttt{tiny} detector consists of a transformer encoder with $6$ layers. 
Each layer consists of a multi-head self-attention module and a feed-forward network (FFN). Each image patch is linear-mapped to a $192$-dimensional representation vector. 
The projector outputs single-scale feature maps with $256$ channels. There are $100$ object queries for the decoder.

The \texttt{small} detector contains $10$ encoder layers, and $300$ object queries. Same as the \texttt{tiny} detector, the dimensions for the input patch representation and the output of the projector are $192$ and $256$. The \texttt{medium} detector is similar to \texttt{small}, 
and the differences include that the dimension of the input patch representation is $384$,
and accordingly the dimension for the encoder is $384$. 

The \texttt{large} detector consists of a $10$-layer encoder and uses two-scale feature maps (see the part \textbf{Projector} in Section~\ref{sec:LW-DETR:Architecture}). The dimensions for the input patch representation and the output of the projector are $384$ and $384$. The \texttt{xlarge} detector is similar to \texttt{large}, and the difference is that the dimension of the input patch representation is $768$. 

\begin{table}[t]
  \centering
    \setlength{\tabcolsep}{6pt}
    \renewcommand{\arraystretch}{1.41}
    \tiny
  \caption{\textbf{Architectures of five LW-DETR instances.}
  }
  \vspace{-2mm}
  \begin{tabular}{l|cccc|ccc|cc}
    & \multicolumn{4}{c|}{ViT encoder} & \multicolumn{3}{c|}{Projector} & \multicolumn{2}{c}{DETR decoder} \\
    LW-DETR &  \#layers & dim & \makecell{\#global\\ attention} & \makecell{\#window\\attention} & \#blocks & dim & scales & \#layers  & \makecell{\#object\\queries}\\
    \shline
    \texttt{tiny} & 6 & 192 & 3&3 & $ 1$ & 256 & $\frac{1}{16}$ & 3 &  100 \\
    \texttt{small}  & 10 & 192 & 4&6 & $ 1$ & 256 & $\frac{1}{16}$ & 3 &  300 \\
    \texttt{medium}  & 10 & 384 & 4&6 & $ 1$ & 256 & $\frac{1}{16}$ & 3 &  300 \\
    \texttt{large} & 10 & 384 & 4&6 & $ 1$ & 
 384 & $\frac{1}{8}, \frac{1}{32}$ & 3 &  300 \\
    \texttt{xlarge}  & 10 & 768 & 4&6 & $ 1$ & 384 & $\frac{1}{8}, \frac{1}{32}$ & 3  & 300 \\
  \end{tabular}
  \label{tab:modelscaling}
  \vspace{-2mm}
\end{table}

\subsection{Effective Training}
\noindent\textbf{More supervision.} Various techniques have been developed to introduce more supervision for accelerating the DETR training, e.g.,~\cite{chen2023group,jia2023detrs,zong2023detrs}.
We adopt Group DETR~\cite{chen2023group} that is easily implemented and 
does not change the inference process. Following~\cite{chen2023group}, we use $13$ parallel weight-sharing decoders for training. For each decoder, we generate the object queries for each group from the output features of the projector.
Following~\cite{chen2023group}, we use the primary decoder 
for the inference.

\vspace{1mm}
\noindent\textbf{Pretraining on Objects365.} The pretraining process consists of two stages. First, we pretrain the ViT on the dataset Objects365 using a MIM method, CAEv2~\cite{zhang2023cae}, based on the pretrained models. This leads to a $0.7$ mAP gain on COCO. 

Second, we follow~\cite{zhang2022dino,chen2022group} to retrain the encoder and train the projector and the decoder on Objects365
in a supervision manner. 

\subsection{Efficient Inference} \label{sec:efficient_inference}
We make a simple modification~\cite{li2021benchmarking,li2022exploring} and adopt interleaved window and global attentions: replacing some global self-attention layers with window self-attention layers. For example, in a $6$-layer ViT, the first, third, and fifth layers are implemented with window attentions. The window attention is implemented by partitioning the feature map into non-overlapping windows and performing self-attention over each window separately.

We adopt a {\em window-major feature map organization} scheme
for efficient interleaved attention, which organizes the feature maps window by window. The ViTDet implementation~\cite{li2022exploring}, where feature maps are organized row by row (row-major organization), requires costly $\operatorname{permutation}$ operations to transition feature maps from a row-major to a window-major organization for window attention. 
Our implementation removes these operations and thus reduces model latency.

We illustrate the window-major way using a toy example. 
Given a $4\times 4$ feature map
\begin{align}
\begin{bmatrix}
f_{11} & f_{12} & f_{13} & f_{14}\\
f_{21} & f_{22} & f_{23} & f_{24}\\
f_{31} & f_{32} & f_{33} & f_{34}\\
f_{41} & f_{42} & f_{43} & f_{44}\\
\end{bmatrix},
\end{align}
the window-major organization for a window size $2 \times 2$
is as below:
\begin{equation}
\begin{aligned}
    &f_{11}, f_{12}, f_{21}, f_{22};
    f_{13}, f_{14}, f_{23}, f_{24};\\
    &f_{31}, f_{32}, f_{41}, f_{42}; 
    f_{33}, f_{34}, f_{43}, f_{44}.
\end{aligned}
\end{equation}
This organization is applicable to both window attention and global attention without rearranging the features. The row-major organization,
\begin{equation}
\begin{aligned}
&f_{11}, f_{12}, f_{13}, f_{14};
f_{21}, f_{22}, f_{23}, f_{24};\\
&f_{31}, f_{32}, f_{33}, f_{34};
f_{41}, f_{42}, f_{43}, f_{44},
\end{aligned}
\end{equation}
is fine for global attention, and needs to be processed with the costly $\operatorname{permutation}$ operation for performing window attention. 

\begin{table}[t]
  \centering
    \setlength{\tabcolsep}{8pt}
    \renewcommand{\arraystretch}{1.41}
    \tiny
    \caption{\textbf{The influence of
    effective training and efficient inference techniques.} We show empirical results from an initial detector with global attention layers to the final LW-DETR-\texttt{small} model. `$\dagger$' means we use the ViTDet implementation. The results except the last row are obtained under $45$K iterations
    (equal to $12$ epochs). 
    The last row corresponds to the result of the final model with $180$K training iterations.}
    \vspace{-2mm}
  \begin{tabular}{l|ccc|c}
    model settings & \#Params (M) & FLOPs (G) & Latency (ms) & mAP \\
    \shline
    initial detector & 10.8 & 22.8 & 3.6 & 35.3 \\
    \hline
    + multi-level feature aggregation & 11.0 & 23.0 & 3.7 & 36.0 \\
    + interleaved window and global attention$^\dagger$ & 11.0 & 16.6 & 3.9 & 34.7 \\
    + window-major feature map organization & 11.0 & 16.6 & 2.9 & 34.7 \\
    \hline
    + iou-aware classification loss & 11.0 & 16.6 & 2.9 & 35.4 \\
    + more supervision & 14.6 & 16.6 & 2.9 & 38.4 \\
    + bounding box reparameterization & 14.6 & 16.6 & 2.9 & 38.6 \\
    + pretraining on Objects365 & 14.6 & 16.6 & 2.9 & 47.3 \\
    \hline
    LW-DETR-\texttt{small} & 14.6 & 16.6 & 2.9 & 48.0 \\
  \end{tabular}
  \label{tab:empirical_study}
  \vspace{-3mm}
\end{table}

\subsection{Empirical Study}~\label{sec:empirical_study}
We empirically show how effective training and efficient inference techniques improve the DETR. 
We use the \texttt{small} detector as the example. The study is based on an initial detector: the encoder is formed with global attention in all the layers and outputs the feature map of the last layer. The results are shown in Table~\ref{tab:empirical_study}.

\vspace{1mm}
\noindent\textbf{Latency improvements.} The interleaved window and global attention, adopted by ViTDet, reduces the computational complexity from $23.0$ GFlops to $16.6$ GFlops, validating the benefit of replacing expensive global attention with cheaper window attention. The latency is not reduced and even increased by $0.2$ ms. This is because extra costly $\operatorname{permutation}$ operations are needed in the row-major feature map organization. Window-major feature map organization alleviates the side effects and leads to a larger latency reduction of $0.8$ ms, from $3.7$ ms to $2.9$ ms.

\vspace{1mm}
\noindent\textbf{Performance improvements.} Multi-level feature aggregation brings a $0.7$ mAP gain. Iou-aware classification loss and more supervision improve the mAP scores from $34.7$ to $35.4$ and $38.4$. Bounding box reparameterization for box regression target~\cite{lin2023detr} (details in the Supplementary Material) makes a slight performance improvement. The significant improvement comes from pretraining on Objects365 and reaches $8.7$ mAP, implying that the transformer indeed benefits from large data. A longer training schedule can give further improvements, forming our LW-DETR-\texttt{small} model.

\section{Experiments}
\subsection{Settings}
\noindent\textbf{Datasets.} 
The dataset for pretraining is Objects365~\cite{shao2019objects365}. We follow~\cite{zhang2022dino,chen2022group} to combine the images of the \texttt{train} set and the images in the \texttt{validate} set except the first $5k$ images for detection pretraining. We use the standard COCO2017~\cite{lin2014microsoft} data splitting policy and perform the evaluation on COCO \texttt{val2017}.

\vspace{1mm}
\noindent\textbf{Data augmentations.} 
We adopt the data augmentations in the DETR and its variants~\cite{carion2020end,zhu2020deformable}. We follow the real-time detection algorithms~\cite{supergradients,yolov8_ultralytics,lyu2022rtmdet} and randomly resize the images into squares for training. For evaluating the performance and the inference time, we follow the evaluation scheme used in the real-time detection algorithms~\cite{supergradients,yolov8_ultralytics,lyu2022rtmdet} to resize the images to $640\times 640$. We use a window size of $10\times 10$ to make sure that the image size can be divisible by the window size.

\vspace{1mm}
\noindent\textbf{Implementation details.} We pretrain the detection model on Objects365~\cite{shao2019objects365} for $30$ epochs and finetune the model on COCO~\cite{lin2014microsoft} for a total number of $180$K training iterations. We adopt the exponential moving average (EMA) technique~\cite{tarvainen2017mean} with a decay of $0.9997$. We use the AdamW optimizer~\cite{loshchilov2017decoupled} for training.

For pretraining, we set the initial learning rate of the projector and the DETR decoder as $4\times e^{-4}$, the initial learning rate of the ViT backbone is $6\times e^{-4}$, and the batch size is $128$. For fine-tuning, we set the initial learning rate of the projector and the DETR decoder as $1\times e^{-4}$, and the initial learning rate of the ViT backbone as $1.5\times e^{-4}$. We set the batch size as $32$ in the \texttt{tiny}, \texttt{small}, and \texttt{medium} models, and the batch size as $16$ in the \texttt{large} and \texttt{xlarge} models. The number of training iterations $180$K is $50$ epochs for the \texttt{tiny}, \texttt{small}, and \texttt{medium} models, and $25$ epochs for the \texttt{large} and \texttt{xlarge} models. More details, such as weight decay, layer-wise decay in the ViT encoder, and component-wise decay~\cite{chen2022group} in the fine-tuning process, are given in the Supplementary Material. 

We measure the averaged inference latency in an end-to-end manner with \texttt{fp16} precision and a batch size of 1 on COCO \texttt{val2017} with a T4 GPU, where the environment settings are with~\texttt{TensorRT-8.6.1},~\texttt{CUDA-11.6}, and~\texttt{CuDNN-8.7.0}. The~\texttt{efficientNMSPlugin} in TensorRT is adopted for efficient NMS implementation. The performance and the end-to-end latency are measured for all real-time detectors using the official implementations.

\begin{table}[t]
  \centering
    \setlength{\tabcolsep}{4pt}
    \renewcommand{\arraystretch}{1.41}
    \tiny
  \caption{\textbf{Comparisons with state-of-the-art real-time detectors}, including RTMDet~\cite{lyu2022rtmdet}, YOLOv8~\cite{yolov8_ultralytics}, and YOLO-NAS~\cite{supergradients}. 
  The total latency is evaluated in an end-to-end manner on COCO \texttt{val2017} and 
  includes the model latency and 
  the postprocessing procedure NMS for non-DETR methods. 
  We measure the total latency in two settings for NMS: official implementation and tuned score threshold. 
  Our LW-DETR does not need NMS and 
  the total latency is equal to 
  the model latency. 
  `pretraining' means the result is based on pretraining on Objects365.
  }
  \vspace{-2mm}
    \begin{tabular}{l|c|ccc|cc|cc}
   \multirow{2}{*}{Method} & \multirow{2}{*}{pretraining} & \multirow{2}{*}{\makecell{\#Params\\(M)}} & \multirow{2}{*}{\makecell{FLOPs\\(G)}} & \multirow{2}{*}{\makecell{Model\\Latency\\(ms)}} & \multicolumn{2}{c|}{official implementation} & \multicolumn{2}{c}{tuned score threshold} \\
     & & & & & \makecell{Total Latency\\(ms)} & mAP & \makecell{Total Latency\\(ms)} & mAP\\
    \shline
    RTMDet-tiny &  & 4.9 & 8.1 & 2.1 & 7.4 & 41.0 & 2.4 & 40.8 \\
    RTMDet-tiny & \checkmark & 4.9 & 8.1 & 2.1 & 7.4 & 41.7 & 2.4 & 41.5 \\
    YOLOv8n & & 3.2 & 4.4 & 1.5 & 6.2 & 37.4 & 1.6 & 37.3 \\
    YOLOv8n & \checkmark & 3.2 & 4.4 & 1.5 & 6.2 & 37.6 & 1.6 & 37.5 \\
    \rowcolor{lightgray!25} LW-DETR-\texttt{tiny} & \checkmark & 12.1 & 11.2 & 2.0 & 2.0 & 42.6 & - & - \\
    \shline
    RTMDet-s & & 8.9 & 14.8 & 2.8 & 7.9 & 44.6 & 2.9 & 44.4 \\
    RTMDet-s & \checkmark & 8.9 & 14.8 & 2.8 & 7.9 & 44.9 & 2.9 & 44.7 \\
    YOLOv8s & & 11.2 & 14.4 & 2.6 & 7.0 & 45.0 & 2.7 & 44.8 \\
    YOLOv8s & \checkmark & 11.2 & 14.4 & 2.6 & 7.0 & 45.2 & 2.7 & 45.1 \\
    YOLO-NAS-s & \checkmark & 19.0 & 17.6 & 2.8 & 4.7 & 47.6 & 2.9 & 47.3 \\
    \rowcolor{lightgray!25} LW-DETR-\texttt{small} & \checkmark & 14.6 & 16.6 & 2.9 & 2.9 & 48.0 & - & - \\
    \shline
    RTMDet-m & & 24.7 & 39.2 & 6.2 & 10.8 & 49.3 & 6.5 & 49.1 \\
    RTMDet-m & \checkmark & 24.7 & 39.2 & 6.2 & 10.8 & 49.7 & 6.5 & 49.5 \\
    YOLOv8m & & 25.6 & 39.7 & 5.9 & 10.1 & 50.3 & 6.0 & 50.0 \\
    YOLOv8m & \checkmark & 25.6 & 39.7 & 5.9 & 10.1 & 50.6 & 6.0 & 50.4 \\
    YOLO-NAS-m & \checkmark & 51.1 & 48.0 & 5.5 & 7.8 & 51.6 & 5.7 & 51.1 \\
    \rowcolor{lightgray!25} LW-DETR-\texttt{medium} & \checkmark & 28.2 & 42.8 & 5.6 & 5.6 & 52.5 & - & - \\
    \shline
    RTMDet-l & & 52.3 & 80.1 & 10.3 & 14.9 & 51.4 & 10.5 & 51.2 \\
    RTMDet-l & \checkmark & 52.3 & 80.1 & 10.3 & 14.9 & 52.4 & 10.5 & 52.2 \\
    YOLOv8l & & 43.7 & 82.7 & 9.3 & 13.2 & 53.0 & 9.4 & 52.5 \\
    YOLOv8l & \checkmark & 43.7 & 82.7 & 9.3 & 13.2 & 53.3 & 9.4 & 53.0 \\
    YOLO-NAS-l & \checkmark & 66.9 & 65.5 & 7.5 & 8.8 & 52.3 & 7.6 & 51.9 \\
    \rowcolor{lightgray!25} LW-DETR-\texttt{large} & \checkmark & 46.8 & 71.6 & 8.8 & 8.8 & 56.1 & - & - \\
    \shline
    RTMDet-x & & 94.9 & 141.7 & 18.4 & 22.8 & 52.8 & 18.8 & 52.5 \\
    RTMDet-x & \checkmark & 94.9 & 141.7 & 18.4  & 22.8 & 54.0 & 18.8 & 53.5 \\
    YOLOv8x & & 68.2 & 129.3 & 14.8 & 19.1 & 54.0 & 15.0 & 53.5 \\
    YOLOv8x & \checkmark & 68.2 & 129.3 & 14.8 & 19.1 & 54.5 & 15.0 & 54.1 \\
    \rowcolor{lightgray!25} LW-DETR-\texttt{xlarge} & \checkmark & 118.0 & 174.2 & 19.1 & 19.1 & 58.3 & - & - \\
  \end{tabular}
  \label{tab:comparesota}
  \vspace{-5mm}
\end{table}

\subsection{Results}
The results of our five LW-DETR models are reported in Table~\ref{tab:comparesota}. LW-DETR-\texttt{tiny} achieves $42.6$ mAP with $500$ FPS on a T4 GPU. LW-DETR-\texttt{small} and LW-DETR-\texttt{medium} get $48.0$ mAP with over $340$ FPS and $52.5$ mAP with a speed of over $178$ FPS respectively. The \texttt{large} and \texttt{xlarge} models achieve $56.1$ mAP with $113$ FPS, and $58.3$ mAP with $52$ FPS.

\vspace{1mm}
\noindent\textbf{Comparisons with state-of-the-art real-time detectors.} 
In Table~\ref{tab:comparesota}, 
we report the comparison of the LW-DETR models against representative real-time detectors, including YOLO-NAS~\cite{supergradients}, YOLOv8~\cite{yolov8_ultralytics}, and RTMDet~\cite{lyu2022rtmdet}. 
One can see that LW-DETR consistently outperforms previous SoTA real-time detectors
with and without using pretraining. 
Our LW-DETR shows clear superiority over YOLOv8 and RTMDet
in terms of latency and detection performance for the five scales from \texttt{tiny} to \texttt{xlarge}. 

In comparison to one of previous best methods YOLO-NAS,
that is obtained with neural architecture search,
our LW-DETR model outperforms it by $0.4$ mAP and $0.9$ mAP, and runs $1.6\times$ and $\sim$$1.4\times$ faster at the \texttt{small} and \texttt{medium} scales. When the model gets larger, the improvement becomes more significant: a $3.8$ mAP improvement when running at the same speed 
at the \texttt{large} scale. 

We further improve other methods
by well-tuning the classification score threshold in the NMS procedure,
and report the results 
in the right two columns.
The results are greatly improved,
and still lower than our LW-DETR.
We expect that our approach potentially
benefits
from other improvements, such as 
neural architecture search (NAS), 
data augmentation, pseudo-labeled data, and knowledge distillation that are exploited
by previous real-time  detectors~\cite{lyu2022rtmdet,yolov8_ultralytics,supergradients}.

\vspace{1mm}
\noindent\textbf{Comparison with concurrent works.} We compare our LW-DETR with concurrent works in real-time detection, YOLO-MS~\cite{chen2023yolo}, Gold-YOLO~\cite{wang2023gold}, RT-DETR~\cite{lv2023detrs}, and YOLOv10~\cite{wang2024yolov10}. 
YOLO-MS improves the performance by enhancing the multi-scale feature representations. 
Gold-YOLO boosts the multi-scale feature fusion and applies MAE-style pretraining~\cite{he2022masked} to improve the YOLO performance. 
YOLOv10 designs several efficiency and accuracy driven modules to improve the performance.
RT-DETR~\cite{lv2023detrs}, closely related to LW-DETR, is also built on the DETR framework, with many differences from our approach in the backbone, the projector, the decoder, and the training schemes. 

\begin{figure}[t]
\centering
\begin{tikzpicture}[font=\footnotesize]
\begin{axis}[
legend columns=1, 
legend style={at={(0.85,0.6)},anchor=north, font=\tiny, draw=gray!20}, legend cell align={left},
y label style={at={(-0.1,0.5)}},
ylabel={mAP}, xlabel={Inference Time (ms)}, ymajorgrids=true, 
extra y ticks={59.0},
tick style={draw=none},
y label style={font=\footnotesize},
height=7cm,
width=8.5cm,
axis lines = left,
every outer y axis line/.style={draw=gray!40},
every outer x axis line/.style={draw=gray!40},
grid style={line width=.1pt, draw=gray!20},
xmin=0, ymax=59.0, ymin=38.0]
\addplot[line width=1.0pt, mark size=1.5pt, mark=diamond, draw=red, draw opacity=0.8]
table
{
X Y
2.0 42.6
2.9 48.0
5.6 52.5
8.8 56.1
19.1 58.3
};
\addplot[line width=1.0pt, mark size=1.5pt, mark=x, draw=chromeyellow!65!dark, draw opacity=0.8]
table
{
X Y
4.6 49.2
9.3 55.3
13.5 56.2
};

\addplot[line width=1.0pt, mark size=1.5pt, mark=x, draw=mossgreen2!80!dark, draw opacity=0.8]
table
{
X Y
3.2 43.8
5.6 46.7
};
\addplot[line width=1.0pt, mark size=1.5pt, mark=x, draw=babyblue!80!dark, draw opacity=0.8]
table
{
X Y
3.4 46.0
6.1 50.3
};
\addplot[line width=1.0pt, mark size=1.5pt, mark=x, draw=mossgreen2!40!white, draw opacity=0.8]
table
{
X Y
6.9 43.9
9.2 46.8
};
\addplot[line width=1.0pt, mark size=1.5pt, mark=x, draw=babyblue!60!white, draw opacity=0.5]
table
{
X Y
3.6 46.1
6.3 50.4
};

\addplot[line width=1.0pt, mark size=1.5pt, mark=x, draw=red!50!yellow, draw opacity=0.4]
table
{
X Y
1.84 38.5
2.49 46.3
4.74 51.1
5.74 52.5
7.28 53.2
10.70 54.4
};
\addlegendentry{LW-DETR}  
\addlegendentry{RT-DETR} 
\addlegendentry{YOLO-MS*} 
\addlegendentry{Gold-YOLO*}  
\addlegendentry{YOLO-MS} 
\addlegendentry{Gold-YOLO}
\addlegendentry{YOLOv10}
\end{axis}
\end{tikzpicture}
\vspace{-3mm}
\caption{\textbf{Our approach outperforms concurrent works.} The x-axis corresponds to the inference time. The y-axis corresponds to the mAP score on COCO \texttt{val2017}. Our LW-DETR,  RT-DETR~\cite{lv2023detrs}, YOLO-MS~\cite{chen2023yolo}, and Gold-YOLO~\cite{wang2023gold} are trained with pretraining on Objects365, while YOLOv10~\cite{wang2024yolov10} is not. The NMS post-processing times are included for YOLO-MS and Gold-YOLO, and measured on the COCO \texttt{val2017} with the setting from the official implementation, and the well-tuned NMS postprocessing setting (labeled as ``*").}
\label{fig:compareconcurrent}
\vspace{-3mm}
\end{figure}
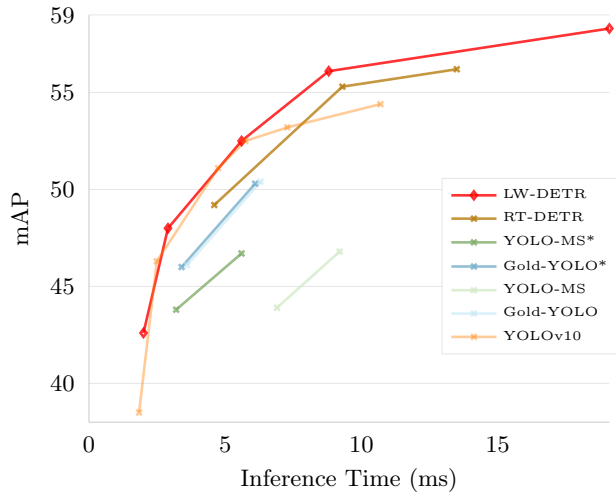

The comparisons are given in Table~\ref{tab:compareconcurrent} and Figure~\ref{fig:compareconcurrent}. 
Our LW-DETR consistently achieves a better balance between the detection performance and the latency. 
YOLO-MS and Gold-YOLO clearly show worse results than our LW-DETR
for all the model scales.
LW-DETR-\texttt{large} outperforms the closely related RT-DETR-R50 by $0.8$ mAP and shows faster speed ($8.8$ ms {vs.} $9.9$ ms). 
LW-DETR with other scales also shows
better results than RT-DETR.
Compared to the latest work, YOLOv10-X~\cite{wang2024yolov10}, our LW-DETR-\texttt{large} achieves higher performance ($56.1$ mAP {vs.} $54.4$ mAP) with lower latency ($8.8$ ms {vs.} $10.70$ ms).

\begin{table}[t]
  \centering
    \setlength{\tabcolsep}{4pt}
    \renewcommand{\arraystretch}{1.41}
    \tiny
  \caption{\textbf{Comparisons with concurrent works}, including YOLO-MS~\cite{chen2023yolo}, Gold-YOLO~\cite{wang2023gold}, YOLOv10~\cite{wang2024yolov10}, and RT-DETR~\cite{lv2023detrs} on COCO. For YOLO-MS and Gold-YOLO, 
  we measure the total latency in two settings for NMS: official implementation and tuned score threshold. 
  For YOLOv10, we report the results in the official paper~\cite{wang2024yolov10}.
  RT-DETR is based on DETR, and the total latency is equal to the model latency. We provide the best latency among the reported inference time in the paper~\cite{lv2023detrs} and the measured time in our environment for RT-DETR. 
  LW-DETR consistently gets superior results. `pretraining' means that the results are based on pretraining on Objects365.
  }
  \vspace{-2mm}
  \begin{tabular}{l|c|ccc|cc|cc}
    \multirow{2}{*}{Method} & \multirow{2}{*}{pretraining} & \multirow{2}{*}{\makecell{\#Params\\(M)}} & \multirow{2}{*}{\makecell{FLOPs\\(G)}} & \multirow{2}{*}{\makecell{Model\\Latency\\(ms)}} & \multicolumn{2}{c|}{official implementation} & \multicolumn{2}{c}{tuned score threshold} \\
     & & & & & \makecell{Total Latency\\(ms)} & mAP & \makecell{Total Latency\\(ms)} & mAP\\
    \shline
    YOLO-MS-XS & & 4.5 & 8.7 & 3.0 & 6.9 & 43.4 & 3.2 & 43.3 \\
    YOLO-MS-XS & \checkmark & 4.5 & 8.7 & 3.0 & 6.9 & 43.9 & 3.2 & 43.8 \\
    YOLO-MS-S & & 8.1 & 15.6 & 5.4 & 9.2 & 46.2 & 5.6 & 46.1 \\
    YOLO-MS-S & \checkmark & 8.1 & 15.6 & 5.4 & 9.2 & 46.8 & 5.6 & 46.7 \\
    YOLO-MS & & 22.0 & 40.1 & 8.6 & 12.3 & 51.0 & 9.0 & 50.8 \\
    \shline
    Gold-YOLO-S & & 21.5 & 23.0 & 2.9 & 3.6 & 45.5 & 3.4 & 45.4 \\
    Gold-YOLO-S & \checkmark & 21.5 & 23.0 & 2.9 & 3.6 & 46.1 & 3.4 & 46.0 \\
    Gold-YOLO-M & & 41.3 & 43.8 & 5.8 & 6.3 & 50.2 & 6.1 &  50.2 \\
    Gold-YOLO-M & \checkmark & 41.3 & 43.8 & 5.8 & 6.3 & 50.4 & 6.1 &  50.3 \\
    Gold-YOLO-L & & 75.1 & 75.9 & 10.2 & 10.6 & 52.3 & 10.5 &  52.2 \\ 
    \shline
    YOLOv10-N & & 2.3 & 6.7 & - & 1.84 & 38.5 & - & - \\
    YOLOv10-S & & 7.2 & 21.6 & - & 2.49 & 46.3 & - & - \\
    YOLOv10-M & & 15.4 & 59.1 & - & 4.74 & 51.1 & - & - \\
    YOLOv10-B & & 19.1 & 92.0 & - & 5.74 & 52.5 & - & - \\
    YOLOv10-L & & 24.4 & 120.3 & - & 7.28 & 53.2 & - & - \\
    YOLOv10-X & & 29.5 & 160.4 & - & 10.70 & 54.4 & - & - \\
    \shline
    RT-DETR-R18 & & 20 & 30.0 & 4.6 & 4.6 & 46.5 & - & - \\
    RT-DETR-R18 & \checkmark & 20 & 30.0 & 4.6 & 4.6 & 49.2 & - & - \\
    RT-DETR-R50 & & 42 & 69.4 & 9.3 & 9.3 & 53.1 & - & - \\
    RT-DETR-R50 & \checkmark & 42 & 69.4 & 9.3 & 9.3 & 55.3 & - & - \\
    RT-DETR-R101 & & 76 & 131.0 & 13.5 & 13.5 & 54.3 & - & - \\
    RT-DETR-R101 & \checkmark & 76 & 131.0 & 13.5 & 13.5 & 56.2 & - & - \\
    \shline
    \rowcolor{lightgray!25} LW-DETR-\texttt{tiny} & \checkmark & 12.1 & 11.2 & 2.0 & 2.0 & 42.6 & - & - \\
    \rowcolor{lightgray!25} LW-DETR-\texttt{small} & \checkmark & 14.6 & 16.6 & 2.9 & 2.9 & 48.0 & - & - \\
    \rowcolor{lightgray!25} LW-DETR-\texttt{medium} & \checkmark & 28.2 & 42.8 & 5.6 & 5.6 & 52.5 & - & - \\
    \rowcolor{lightgray!25} LW-DETR-\texttt{large} & \checkmark & 46.8 & 71.6 & 8.8 & 8.8 & 56.1 & - & - \\
    \rowcolor{lightgray!25} LW-DETR-\texttt{xlarge} & \checkmark & 118.0 & 174.2 & 19.1 & 19.1 & 58.3 & - & - \\
  \end{tabular}
  \label{tab:compareconcurrent}
  \vspace{-5mm}
\end{table}

\subsection{Discussions}
\noindent\textbf{NMS post-processing.} 
The DETR method is an end-to-end algorithm that does not need the NMS post-processing procedure. 
In contrast, existing real-time detectors, such as YOLO-NAS~\cite{supergradients}, YOLOv8~\cite{yolov8_ultralytics}, and RTMDet~\cite{lyu2022rtmdet}, needs NMS~\cite{hosang2017learning} post-processing. 
The NMS procedure takes extra time.
We include the extra time
for measuring the end-to-end inference cost,
which is counted in real-world application.
The results using the NMS setting in
the official implementations are shown in
Figure~\ref{fig:latencyvsperformance} and Table~\ref{tab:comparesota}.

We further make improvements for the methods with NMS 
by tuning the classification score threshold 
for the NMS post-processing. 
We observe that the default score threshold, 
$0.001$, in YOLO-NAS, YOLOv8, and RTMDet, 
results in a high mAP, but a large number of boxes and thus high latency. 
In particular, when the model is small, 
the end-to-end latency is dominated by the NMS latency. We tune the threshold, 
obtaining a good balance between the mAP score and the latency. 
It is observed that the mAP scores are slightly dropped, e.g., by $-0.1$ mAP to $-0.5$ mAP, and the running times are largely reduced, e.g., 
with a reduction of $4$$\sim$$5$ ms for RTMDet and YOLOv8, a reduction of $1$$\sim$$2$ ms for YOLO-NAS. These reductions are from that fewer predicted boxes are fed into NMS after tuning the score threshold. Detailed results with different score thresholds, along with the distribution of the number of remaining boxes across the COCO \texttt{val2017} are given in the Supplementary Material.

Figure~\ref{fig:latencyvsperformance} shows the comparison against other methods with well-tuned NMS procedures.
The methods with NMS are improved. 
Our approach still outperforms other methods. 
The second-best approach, YOLO-NAS, is a network architecture search algorithm and performs very closely to the proposed baseline. We believe that the complicated network architecture search procedure, like the one used in YOLO-NAS, potentially benefits the DETR approach, and further improvement is expected.

\begin{table}[t]
  \centering
    \setlength{\tabcolsep}{6pt}
    \renewcommand{\arraystretch}{1.41}
    \tiny
  \caption{\textbf{
  The effect of pertaining in our LW-DETR.} Pretraining on Objects365 improves our approach a lot. This observation is consistent with the observations from methods with large models~\cite{zhang2022dino,chen2022group,zong2023detrs}.}
  \vspace{-1mm}
  \begin{tabular}{l|ccc|c|c}
    LW-DETR & \#Params (M) & FLOPs (G) & Latency (ms) & mAP w/o pretraining & mAP w/ pretraining \\
    \shline
    \texttt{tiny} & 12.1 & 11.2 & 2.0 & 36.5 & 42.6 \\
    \texttt{small} & 14.6 & 16.6 & 2.9 & 43.6 & 48.0 \\
    \texttt{medium} & 28.2 & 42.8 & 5.6 & 47.2 & 52.5 \\
    \texttt{large} & 46.8 & 71.6 & 8.8 & 49.5 & 56.1 \\
    \texttt{xlarge} & 118.0 & 174.2 & 19.1 & 53.0 & 58.3 \\
    \shline
    R18 & 21.2 & 21.4 & 2.5 & 40.9 & 44.4 \\
    R50 & 54.6 & 67.7 & 8.7 & 49.7 & 54.4 \\
  \end{tabular}
  \label{tab:LW-DETRpretrain}
  \vspace{-6mm}
 \end{table}

\begin{table}[t]
  \centering
    \setlength{\tabcolsep}{4.5pt}
    \renewcommand{\arraystretch}{1.41}
    \tiny
  \caption{\textbf{
  The effect of pertaining along with training epochs for non-end-to-end detectors.}}
  \vspace{-1mm}
  \begin{tabular}{l|c|c|c|c|c|c|c}
    Model & pretraining & 20 epochs & 60 epochs & 100 epochs & 200 epochs & 300 epochs & 500 epochs \\
    \shline
    YOLOv8n & & 26.2 & 30.1 & 32.3 & 34.0 & 35.0 & 37.4 \\
    YOLOv8n & \checkmark & 31.5 & 32.8 & 33.2 & 34.3 & 35.2 & 37.6 \\
    \hline
    RTMDet-t & & 30.4 & 34.2 & 35.2 & 36.8 & 41.0 & -  \\
    RTMDet-t & \checkmark & 33.4 & 36.5 & 36.7 & 37.5 & 41.7 & - \\
    \hline
    YOLO-MS-XS & & 24.7 & 34.3 & 36.4 & 39.2 & 43.4 & -  \\
    YOLO-MS-XS & \checkmark & 37.5 & 38.6 & 39.1 & 40.0 & 43.9 & -  \\
    \hline
    Gold-YOLO-S & & 33.4 & 37.1 & 38.2 & 43.6 & 45.5 & - \\
    Gold-YOLO-S & \checkmark & 39.3 & 41.3 & 42.1 & 44.9 & 46.1 & - \\
  \end{tabular}
  \label{tab:gainalongepochs}
  \vspace{-2mm}
 \end{table}

\noindent\textbf{Pretraining.} We empirically study the effect of pretraining. The results, shown in Table~\ref{tab:LW-DETRpretrain}, indicate that pretraining leads to significant improvements for our approaches, with an average improvement of $5.5$ mAP. The \texttt{tiny} model gets an mAP gain of $6.1$, and the \texttt{xlarge} model gets an mAP gain of $5.3$. This implies that pretraining on a large dataset is highly beneficial for DETR-based models.

We further show that the training procedure applies to the DETR approach with convolutional encoders. We replace transformer encoders with ResNet-$18$ and ResNet-$50$. One can see that in 
Table~\ref{tab:LW-DETRpretrain}, 
the results of these LW-DETR variants are close to LW-DETR with transformer encoders in terms of latency and mAP, 
and the pretraining brings benefits
that are similar to and a little lower than 
LW-DETR with transformer encoders.

Meanwhile, we investigate the pretraining improvements on non-end-to-end detectors. 
According to the results in Table~\ref{tab:comparesota}, Table~\ref{tab:compareconcurrent} and Table~\ref{tab:gainalongepochs}, it seems that pretraining on Objects365 only show limited gains for non-end-to-end detectors~\cite{lyu2022rtmdet,yolov8_ultralytics,chen2023yolo,wang2023gold}, which is different from the phenomenon in DETR-based detectors, where pretraining give large improvements. As the non-end-to-end detectors train $300$ epochs even $500$ epochs in YOLOv8, we wonder if the limited gain is related to the training epochs. We compare the improvements brought by the pretrained weights along with the training epochs. Table~\ref{tab:gainalongepochs} shows that the improvements diminished along with the training epochs, which partly supports the above hypothesis. The above illustration is a preliminary step. We believe that more investigations are needed to figure out the underlying reasons for the difference in benefits of pretraining.

\subsection{Experiments on more datasets}
We test the generalizability of our LW-DETR on more detection datasets. We consider two types of evaluation methods, cross-domain evaluation and multi-domain finetuning. For cross-domain evaluation, we directly evaluate the real-time detectors trained on COCO on the Unidentified Video Objects (UVO)~\cite{wang2021unidentified}. For multi-domain fine-tuning, we finetune the pretrained real-time detectors on the multi-domain detection dataset Roboflow 100 (RF100)~\cite{ciaglia2022roboflow}. We do a coarse search on the hyperparameters on each dataset for all models, such as the learning rate. Please refer to the Supplementary Material for more details.

\begin{table}[t]
  \centering
    \setlength{\tabcolsep}{14pt}
    \renewcommand{\arraystretch}{1.41}
    \tiny
  \caption{\textbf{Cross-domain evaluation on UVO.} We evaluate the performance in a class-agnostic way as UVO is class-agnostic. LW-DETR demonstrates higher AP and AR than other detectors.}
  \vspace{-1mm}
    \begin{tabular}{l|cc|cccc}
    Method & mAP & AP50 & AR@100 & AR$_s$ & AR$_m$ & AR$_l$ \\
    \shline
    RTMDet-s & 29.7 & 43.3 & 55.7 & 26.5 & 49.4 & 71.5\\
    YOLOv8-s & 29.1 & 42.4 & 54.3 & 27.4  & 48.6 & 68.8\\
    YOLO-NAS-s & 31.0 & 44.5 & 55.1 & 25.8  & 48.1 & 71.6 \\
    LW-DETR-\texttt{small} & \textbf{32.3} & \textbf{45.1} & \textbf{59.8} & \textbf{29.4} & \textbf{52.4} & \textbf{77.1} \\
  \end{tabular}
  \label{tab:comparisonsonuvo}
  \vspace{-5mm}
\end{table}

\begin{table}[t]
  \centering
    \setlength{\tabcolsep}{2.5pt}
    \renewcommand{\arraystretch}{1.41}
    \tiny
  \caption{\textbf{Multi-domain finetuning on RF100.} We compare our LW-DETR with previous real-time detectors, including YOLOv5, YOLOv7, RTMDet, YOLOv8, and YOLO-NAS on all data domains of RF100. \textcolor{gray!50}{Gray} entries are the results from the RF100 paper~\cite{ciaglia2022roboflow}. The data domains are aerial, videogames, microscopic, underwater, documents, electromagnetic, and real world. The AP50 metric is used. `-' means that YOLO-NAS~\cite{supergradients} does not report their detailed results.}
  \vspace{-1mm}
  \begin{tabular}{l|c|ccccccc}
    \multirow{2}{*}{Method} & \multirow{2}{*}{Average} & \multicolumn{7}{c}{Domains in Roboflow 100}\\
    & & aerial & videogames & microscopic & underwater & documents & electromagnetic & real world \\
    \shline
    \textcolor{gray!50}{YOLOv5-s} & \textcolor{gray!50}{73.4} & \textcolor{gray!50}{63.6} & \textcolor{gray!50}{85.9} & \textcolor{gray!50}{65.0} & \textcolor{gray!50}{56.0} & \textcolor{gray!50}{71.6} & \textcolor{gray!50}{74.2} & \textcolor{gray!50}{76.9} \\
    \textcolor{gray!50}{YOLOv7-s} & \textcolor{gray!50}{67.4} & \textcolor{gray!50}{50.4} & \textcolor{gray!50}{79.6} & \textcolor{gray!50}{59.1} & \textcolor{gray!50}{66.2} & \textcolor{gray!50}{72.2} & \textcolor{gray!50}{63.9} & \textcolor{gray!50}{70.5} \\
    \hline
    RTMDet-s & 79.2 & 70.1 & 88.0 & 68.1 & 68.0 & 81.0 & 77.4 & 83.3 \\
    YOLOv8-s & 80.1 & 70.7 & 87.9 & 74.7 & 70.2 & 79.8 & 79.0 & 82.9 \\
    YOLO-NAS-s & 81.5 & - & - & - & - & - & - & - \\
    YOLO-NAS-m & 81.8 & - & - & - & - & - & - & - \\
    \hline
    LW-DETR-\texttt{small} & \textbf{82.5} & 71.8 & 88.9 & 74.5 & 69.6 & 86.7 & 84.6 & 85.3 \\
    LW-DETR-\texttt{medium} & \textbf{83.5} & 72.9 & 90.8 & 75.4 & 70.5 & 86.1 & 86.2 & 86.4 \\
  \end{tabular}
  \label{tab:comparisonsonroboflow}
  \vspace{-3mm}
\end{table}

\vspace{1mm}
\noindent\textbf{Cross-domain evaluation.} One possible way to evaluate the generalizability of the models is to directly evaluate them on the datasets with different domains. We adopt a class-agnostic object detection benchmark, UVO~\cite{wang2021unidentified}, where $57\%$ object instances do not belong to any of the $80$ COCO classes. UVO is based on YouTube videos, whose appearance is very different from COCO, e.g., some videos are in egocentric views and have significant motion blur. We evaluate the models trained with COCO (taken from Table~\ref{tab:comparesota}) on the validation split of UVO.

Table~\ref{tab:comparisonsonuvo} provides the results. LW-DETR excels over competing SoTA real-time detectors. Specifically, LW-DETR-\texttt{small} is $1.3$ mAP and $4.1$ AR higher than the best result among RTMDet-s, YOLOv8-s, and YOLO-NAS-s. In terms of recall, it also shows enhanced abilities to detect more objects across different scales: small, medium, and large. The above findings imply that the superiority of our LW-DETR over previous real-time detectors is attributed not to specific tuning for COCO, but to its capacity for producing more generalizable models.

\vspace{1mm}
\noindent\textbf{Multi-domain finetuning.} 
Another way is to finetune the pretrained detectors on small datasets across different domains. RF100 consists of $100$ small datasets, $7$ imagery domains,
$224$k images, and $829$ class labels. It can help researchers test the model's generalizability with real-life data. We finetune the real-time detectors on each small dataset of RF100. 

The results are given in Table~\ref{tab:comparisonsonroboflow}.
LW-DETR-\texttt{small} shows superiority over current state-of-the-art real-time detectors across different domains. In particular, for the `documents' and the `electromagnetic' domains, our LW-DETR is significantly better than YOLOv5, YOLOv7, RTMDet, and YOLOv8 ($5.7$ AP and $5.6$ AP higher than the best among the four). LW-DETR-\texttt{medium} can give further improvements overall. These findings highlight the versatility of our LW-DETR, positioning it as a strong baseline in a range of closed-domain tasks.

\section{Limitation and future works}
Currently, we only demonstrate the effectiveness of LW-DETR in real-time detection. This is the first step. Extending LW-DETR for open-world detection and applying LW-DETR to more vision tasks, such as multi-person pose estimation and multi-view 3D object detection, needs more investigation. We leave them for future work.

\section{Conclusion}
This paper shows that detection transformers 
achieve competitive 
and even
superior results over existing real-time detectors. Our method is simple and efficient. The success stems from
multi-level feature aggregation
and training-effective and inference-efficient techniques. 
We hope our experience can provide insights for building real-time models with transformers in vision tasks.

\bibliographystyle{splncs04}
\bibliography{main}

\clearpage

\renewcommand{\thesection}{\Alph{section}}
\setcounter{section}{0}
\section*{Supplementary Material}
\section{Experimental Details}
This section includes details on the hyper-parameters of pretraining on Objects365~\cite{shao2019objects365}, finetuning on COCO~\cite{lin2014microsoft}, and finetuning on Roboflow 100~\cite{ciaglia2022roboflow}, on the architectures of convolutional encoders, and on the modeling of box regression. We represent the \texttt{tiny}/\texttt{small}/\texttt{medium}/\texttt{large}/\texttt{xlarge} versions of our LW-DETR with T/S/M/L/X in the tables for neat representations.

\subsection{Experimental settings}
\noindent\textbf{Pretraining settings.} The default settings are in Table~\ref{tab:pretrainingsettings}. We do not use the learning rate drop schedule and keep the initial learning rate along with the training process. When performing window attention in the ViT encoder, we fix the number of windows as 16 for different image resolutions for easy implementation. We use layer-wise lr decay~\cite{clark2020electra} following previous MIM methods~\cite{he2022masked,chen2022context,zhang2023cae}.

\begin{table}
  \centering
    \setlength{\tabcolsep}{3.2pt}
    \renewcommand{\arraystretch}{1.41}
    \tiny
  \caption{\textbf{Pretraining settings.}}
  \vspace{-3mm}
  \begin{tabular}{l|l}
    Setting  & Value\\
    \shline
    optimizer & AdamW \\
    base learning rate & $4.0\times e^{-4}$ \\
    encoder learning rate & $6.0\times e^{-4}$ \\
    weight decay & $1\times e^{-4}$ \\
    batch size & 128 \\
    epochs & 30 \\
    training images resolutions & [448, 512, 576, 640, 704, 768, 832, 896] \\
    encoder layer-wise lr decay & 0.8 (T/S), 0.7 (M/L), 0.75 (X) \\
    number of encoder layers & 6 (T), 10(S/M/L/X) \\
    drop path & 0 (T/S/M), 0.05 (L/X) \\
    window numbers & 16 \\
    window attention indexes & [0, 2, 4] (T), [0, 1, 3, 6, 7, 9] (S/M/L/X) \\
    output feature indexes & [0, 2, 4] (T), [2, 4, 5, 9] (S/M/L/X) \\
    feature scales & $\frac{1}{16}$ (T/S/M), [$\frac{1}{8}$, $\frac{1}{32}$] (L/X) \\
    number of object queries & 100 (T), 300 (S/M/L/X) \\
    number of decoder layers & 3 \\
    hidden dimensions & 256 (T/S/M), 384 (L/X) \\
    decoder self-attention heads & 8 (T/S/M), 12 (L/X) \\
    decoder cross-attention heads & 16 (T/S/M), 24 (L/X) \\
    decoder sampling points & 2 (T/S/M), 4 (L/X) \\
    group detr & 13 \\
    ema decay & 0.997
  \end{tabular}
  \label{tab:pretrainingsettings}
  \vspace{-3mm}
\end{table}

\begin{table}
  \centering
    \setlength{\tabcolsep}{11.8pt}
    \renewcommand{\arraystretch}{1.41}
    \tiny
  \caption{\textbf{COCO experimental settings.}}
  \vspace{-3mm}
  \begin{tabular}{l|l}
    Setting  & Value\\
    \shline
    base learning rate & $1.0\times e^{-4}$ \\
    encoder learning rate & $1.5\times e^{-4}$ \\
    weight decay & $1\times e^{-4}$ (T/S/M/L), $1\times e^{-3}$ (X) \\
    batch size & 32 (T/S/M), 16 (L/X) \\
    epochs & 50 (T/S/M), 25 (L/X) \\
    drop path & 0 (T/S/M), 0.1 (L/X) \\
    component-wise lr decay & 0.7 (T/S/M), 0.5 (L/X)
  \end{tabular}
  \label{tab:cocosettings}
  \vspace{-3mm}
\end{table}
\begin{table}
  \centering
    \setlength{\tabcolsep}{11.8pt}
    \renewcommand{\arraystretch}{1.41}
    \tiny
  \caption{\textbf{Roboflow 100 experimental settings.}}
  \vspace{-3mm}
  \begin{tabular}{l|l}
    Setting  & Value\\
    \shline
    base learning rate & $8.0\times e^{-4}$ (S), $3.0\times e^{-4}$ (M) \\
    batch size & 16 (S/M) \\
    encoder learning rate & $1.2\times e^{-3}$ (S), $4.5\times e^{-4}$ (M) \\
    encoder layer-wise lr decay & 0.9 (S), 0.8 (M) \\
    component-wise lr decay & 0.7 (S), 0.9 (M)
  \end{tabular}
  \label{tab:rf100settings}
  \vspace{-5mm}
\end{table}
\vspace{1mm}
\noindent\textbf{COCO experimental settings.} Most of the settings follow the ones in the pretraining stage. We share the modifications of settings in Table~\ref{tab:cocosettings}. When finetuning LW-DETR on COCO, we use component-wise lr decay~\cite{chen2022group}, which gives different scale factors for the learning rate in the ViT encoder, the Projector, and the DETR decoder. For example, the component-wise lr decay is $0.7$ means that we set the lr scale factor as $0.7^0$ for the prediction heads, $0.7^1$ for the transformer decoder layers in DETR decoder, $0.7^2$ for the Projector, and $0.7^3$ for the ViT encoder.

\noindent\textbf{Roboflow 100 experimental settings.} Roboflow 100~\cite{ciaglia2022roboflow} consists of $100$ small datasets. We finetune our LW-DETR on these datasets based on the pretrained model on Objects365~\cite{shao2019objects365}. As the training images are insufficient, we set the batch size as $16$ and finetune the model for $100$ epochs following~\cite{ciaglia2022roboflow} on all small datasets to make sure to have sufficient training iterations. 

We tune the learning rate, encoder layer-wise lr decay, and component-wise lr decay (as shown in Table~\ref{tab:rf100settings}), we do a coarse search on the `microscopic' domain and fix these hyper-parameters for other datasets. The other hyper-parameters are kept same with the finetuning experiments on COCO. We also perform the same processes for RTMDet~\cite{lyu2022rtmdet} and YOLOv8~\cite{yolov8_ultralytics} for fair comparisons.

\subsection{Settings for convolutional encoders} 
We also explore convolutional encoders, ResNet-18 and ResNet-50, in our LW-DETR. We load the ImageNet~\cite{deng2009imagenet} pretrained encoder weights from the RT-DETR repo\footnote{\tiny\url{https://github.com/lyuwenyu/RT-DETR/issues/42\#issue-1860463373}}. Instead of directly outputting multi-level feature maps with scales of $[\frac{1}{8}, \frac{1}{16}, \frac{1}{32}]$, we make a simple modification to only output a feature map in $\frac{1}{16}$. We first upsample the feature map in $\frac{1}{32}$ scale to $\frac{1}{16}$, downsample the feature map in $\frac{1}{8}$ to $\frac{1}{16}$, and then concatenates all the feature maps. We add additional convolution layers to reduce the feature dimension to prevent the final concatenated feature map from having extremely large feature dimensions.

\subsection{Box regression target reparameterization}
Box regression target parameterization is a widely used technique in two-stage and one-stage detectors~\cite{ren2015faster,lin2017focal,bochkovskiy2020yolov4,yolov5}, which predicts the parameters for a box transformation that transforms an input proposal into a predicted box. We follow Plain DETR~\cite{lin2023detr} to use this technique in our LW-DETR.

For box regression in the first stage and each decoder layer, we predict four parameters $[\delta_x, \delta_y, \delta_w, \delta_h]$ to a transformation, which transforms a proposal $[p_{c_x}, p_{c_y}, p_w, p_h]$ to a predicted bounding box $[b_{c_x}, b_{c_y}, b_w, b_h]$ by applying:
\begin{equation}
\begin{aligned}
    b_{c_x} = \delta_x * p_w + p_{c_x}, b_{c_y} = \delta_y * p_h + p_{c_y}, \\
    b_w = \exp(\delta_w) * p_w, b_h = \exp(\delta_h) * p_h.
\end{aligned}
\end{equation}
The predicted box $[b_{c_x}, b_{c_y}, b_w, b_h]$ is used for calculating the box regression losses and for output.

\begin{table}[t]
  \centering
    \setlength{\tabcolsep}{4pt}
    \renewcommand{\arraystretch}{1.41}
    \tiny
  \caption{\textbf{Tuning score threshold for non-end-to-end detectors.} We show how the score threshold affects the time of NMS and the detection performance in YOLO-NAS, YOLOv8, RTMDet, YOLO-MS, and Gold-YOLO. We share the detection performance and total latency under three different score thresholds. The first score threshold is the default one in the official implementations. The score threshold in bold represents a good balance between the mAP score and the NMS latency.}
  \begin{tabular}{lc|cc|cc|cc}
    Model & \makecell{Model Latency\\(ms)} & mAP & \makecell{Total Latency\\(ms)} & mAP & \makecell{Total Latency\\(ms)} & mAP & \makecell{Total Latency\\(ms)} \\
    \shline
    \multicolumn{2}{c|}{Thresholds}  & \multicolumn{2}{c|}{\textit{$\operatorname{score}=0.01$}\protect\footnotemark} & \multicolumn{2}{c|}{$\operatorname{\textbf{score}}\textbf{=}\textbf{0.1}$} & \multicolumn{2}{c}{$\operatorname{score}=0.15$} \\
    \hline
    YOLO-NAS-s & 2.75 & 47.6 & 4.68 & 47.3 & 2.88 & 46.7 & 2.82 \\
    YOLO-NAS-m & 5.52 & 51.6 & 7.76 & 51.1 & 5.70 & 50.6 & 5.53 \\
    YOLO-NAS-l & 7.49 & 52.3 & 8.84 & 51.9 & 7.64 & 51.2 & 7.52 \\
    \shline
    \multicolumn{2}{c|}{Thresholds}  & \multicolumn{2}{c|}{\textit{$\operatorname{score}=0.001$}} & \multicolumn{2}{c|}{$\operatorname{\textbf{score}}\textbf{=}\textbf{0.01}$} & \multicolumn{2}{c}{$\operatorname{score}=0.05$} \\
    \hline
    YOLOv8n & 1.51 & 37.4 & 6.21 & 37.3 & 1.62 & 36.0 & 1.54 \\
    YOLOv8s & 2.64 & 45.0 & 7.00 & 44.8 & 2.71 & 43.7 & 2.70 \\
    YOLOv8m & 5.90 & 50.3 & 10.11 & 50.0 & 6.06 & 49.0 & 5.92 \\
    YOLOv8l & 9.30 & 53.0 & 13.16 & 52.5 & 9.44 & 51.3 & 9.31 \\
    YOLOv8x & 14.88 & 54.0 & 19.17 & 53.5 & 15.09 & 52.3 & 14.91 \\
    \shline
    \multicolumn{2}{c|}{Thresholds}  & \multicolumn{2}{c|}{\textit{$\operatorname{score}=0.001$}} & \multicolumn{2}{c|}{$\operatorname{\textbf{score}}\textbf{=}\textbf{0.1}$} & \multicolumn{2}{c}{$\operatorname{score}=0.25$} \\
    \hline
    RTMDet-t & 2.16 & 41.0 & 7.41 & 40.8 & 2.45 & 39.1 & 2.37 \\
    RTMDet-s & 2.88 & 44.6 & 7.88 & 44.4 & 2.94 & 42.6 & 2.93 \\
    RTMDet-m & 6.27 & 49.3 & 10.82 & 49.1 & 6.52 & 47.2 & 6.31 \\
    RTMDet-l & 10.37 & 51.4 & 14.84 & 51.2 & 10.54 & 49.2 & 10.48 \\
    RTMDet-x & 18.44 & 52.8 & 22.81 & 52.5 & 18.88 & 50.5 & 18.69 \\
    \shline
    \multicolumn{2}{c|}{Thresholds}  & \multicolumn{2}{c|}{\textit{$\operatorname{score}=0.001$}} & \multicolumn{2}{c|}{$\operatorname{\textbf{score}}\textbf{=}\textbf{0.1}$} & \multicolumn{2}{c}{$\operatorname{score}=0.25$} \\
    \hline
    YOLO-MS-XS & 3.02 & 43.4 & 6.99 & 43.3 & 3.26 & 41.9 & 3.19 \\
    YOLO-MS-S & 5.40 & 46.2 & 9.18 & 46.1 & 5.62 & 44.2 & 5.56 \\
    YOLO-MS & 8.56 & 51.0 & 12.38 & 50.8 & 9.09 & 48.8 & 8.87  \\
    \shline
    \multicolumn{2}{c|}{Thresholds}  & \multicolumn{2}{c|}{\textit{$\operatorname{score}=0.03$}} & \multicolumn{2}{c|}{$\operatorname{\textbf{score}}\textbf{=}\textbf{0.05}$} & \multicolumn{2}{c}{$\operatorname{score}=0.25$} \\
    \hline
    Gold-YOLO-S & 2.94 & 45.5 & 3.63 & 45.4 & 3.35 & 43.1 & 3.08 \\
    Gold-YOLO-M & 5.84 & 50.2 & 6.34 & 50.2 & 6.14 & 47.6 & 5.98 \\
    Gold-YOLO-L & 10.15 & 52.3 & 10.58 & 52.2 & 10.46 & 50.5 & 10.22 \\
  \end{tabular}
  \label{tab:nms_analysis}
  \vspace{-3mm}
\end{table}

\footnotetext{\footnotesize We have corrected a typo in the main paper regarding YOLO-NAS: the default score threshold should be 0.01, not the value mentioned in L298.}

\section{Analysis on NMS}
\noindent\textbf{Tuning score threshold.} The score threshold in non-end-to-end detectors, decides the number of predicted boxes that are passed to the NMS, which largely affects the NMS latency. Table~\ref{tab:nms_analysis} verifies it with YOLO-NAS, YOLOv8, RTMDet, YOLO-MS, and Gold-YOLO. A large score threshold can largely reduce the overhead of NMS, but bring negative results to detection performance. We optimize the NMS latency by carefully tuning the score thresholds for non-end-to-end detectors, achieving a balance between the detection performance and the total latency. The overhead brought by NMS is significantly reduced to $0.1$$\sim$$0.5$ ms with slight drops in detection performance.

\begin{figure}[h]
\centering
\footnotesize
\begin{tikzpicture}[font=\footnotesize]
\pgfplotsset{compat=1.11,
    /pgfplots/ybar legend/.style={
    /pgfplots/legend image code/.code={%
       \draw[##1,/tikz/.cd,yshift=-0.3em]
        (0cm,0cm) rectangle (7pt,0.8em);},
   },
}
\begin{axis}[
    width=0.55\linewidth, 
    height=0.3\linewidth,
    ybar,
    symbolic x coords={0, 100, 500, 1000, 5000, 30000},
    xtick=data,
    ylabel=number of images,
    ymin=0,
    ymax=4650,
    bar width=22pt,
    legend style={at={(0.02,0.96)},anchor=north west},
    nodes near coords,
    nodes near coords style={font=\tiny},
    nodes near coords greater equal only/.style={
        small value/.style={
            /tikz/coordinate,
        },
        every node near coord/.append style={
            check for small values/.code={
                \begingroup
                \pgfkeys{/pgf/fpu}
                \pgfmathparse{\pgfplotspointmeta<#1}
                \global\let\result=\pgfmathresult
                \endgroup
                \pgfmathfloatcreate{1}{1.0}{0}
                \let\ONE=\pgfmathresult
                \ifx\result\ONE
                    \pgfkeysalso{/pgfplots/small value}
                \fi
            },
            check for small values,
        },
    },
    nodes near coords greater equal only=0.2,
    axis y line=none,
    axis x line =bottom,
    axis line style={-},
    enlarge x limits=0.04,
    tick style={draw=none},
    y post scale=1.0,
    x post scale=0.9,
    legend style={at={(0.5, 1.05)}, anchor=north, legend columns =-1, draw=none, fill=none, font=\tiny},
    ]
    \addplot[bar shift=12pt,fill=babyblue!80!dark, draw=babyblue!80!dark, fill opacity=0.5, draw opacity=0.5] coordinates {(0, 0) (100, 12) (500, 89) (1000, 1265) (5000, 3634) (30000, 0)};
    \end{axis}
\node[below, yshift=-2pt, font=\footnotesize] at (current bounding box.south) {(a) YOLO-NAS-s score threshold $=0.01$};
\end{tikzpicture}
\quad
\begin{tikzpicture}[font=\footnotesize]
\pgfplotsset{compat=1.11,
    /pgfplots/ybar legend/.style={
    /pgfplots/legend image code/.code={%
       \draw[##1,/tikz/.cd,yshift=-0.3em]
        (0cm,0cm) rectangle (7pt,0.8em);},
   },
}
\begin{axis}[
    width=0.55\linewidth, 
    height=0.3\linewidth,
    ybar,
    symbolic x coords={0, 100, 500, 1000, 5000, 30000},
    xtick=data,
    ylabel=number of images,
    ymin=0,
    ymax=4650,
    bar width=22pt,
    legend style={at={(0.02,0.96)},anchor=north west},
    nodes near coords,
    nodes near coords style={font=\tiny},
    nodes near coords greater equal only/.style={
        small value/.style={
            /tikz/coordinate,
        },
        every node near coord/.append style={
            check for small values/.code={
                \begingroup
                \pgfkeys{/pgf/fpu}
                \pgfmathparse{\pgfplotspointmeta<#1}
                \global\let\result=\pgfmathresult
                \endgroup
                \pgfmathfloatcreate{1}{1.0}{0}
                \let\ONE=\pgfmathresult
                \ifx\result\ONE
                    \pgfkeysalso{/pgfplots/small value}
                \fi
            },
            check for small values,
        },
    },
    nodes near coords greater equal only=0.2,
    axis y line=none,
    axis x line =bottom,
    axis line style={-},
    enlarge x limits=0.04,
    tick style={draw=none},
    y post scale=1.0,
    x post scale=0.9,
    legend style={at={(0.5, 1.05)}, anchor=north, legend columns =-1, draw=none, fill=none, font=\tiny},
    ]
    \addplot[bar shift=12pt,fill=mossgreen2!80!white, draw=mossgreen2!80!white, fill opacity=0.5, draw opacity=0.5] coordinates {(0, 1956) (100, 2421) (500, 525) (1000, 98) (5000, 0) (30000, 0)};
    \end{axis}
\node[below, yshift=-2pt, font=\footnotesize] at (current bounding box.south) {(b) YOLO-NAS-s score threshold $=0.1$};
\end{tikzpicture}
\quad
\begin{tikzpicture}[font=\footnotesize]
\pgfplotsset{compat=1.11,
    /pgfplots/ybar legend/.style={
    /pgfplots/legend image code/.code={%
       \draw[##1,/tikz/.cd,yshift=-0.3em]
        (0cm,0cm) rectangle (7pt,0.8em);},
   },
}
\begin{axis}[
    width=0.55\linewidth, 
    height=0.3\linewidth,
    ybar,
    symbolic x coords={0, 100, 500, 1000, 5000, 30000},
    xtick=data,
    ylabel=number of images,
    ymin=0,
    ymax=4650,
    bar width=22pt,
    legend style={at={(0.02,0.96)},anchor=north west},
    nodes near coords,
    nodes near coords style={font=\tiny},
    nodes near coords greater equal only/.style={
        small value/.style={
            /tikz/coordinate,
        },
        every node near coord/.append style={
            check for small values/.code={
                \begingroup
                \pgfkeys{/pgf/fpu}
                \pgfmathparse{\pgfplotspointmeta<#1}
                \global\let\result=\pgfmathresult
                \endgroup
                \pgfmathfloatcreate{1}{1.0}{0}
                \let\ONE=\pgfmathresult
                \ifx\result\ONE
                    \pgfkeysalso{/pgfplots/small value}
                \fi
            },
            check for small values,
        },
    },
    nodes near coords greater equal only=0.2,
    axis y line=none,
    axis x line =bottom,
    axis line style={-},
    enlarge x limits=0.04,
    tick style={draw=none},
    y post scale=1.0,
    x post scale=0.9,
    legend style={at={(0.5, 1.05)}, anchor=north, legend columns =-1, draw=none, fill=none, font=\tiny},
    ]
    \addplot[bar shift=12pt,fill=chromeyellow!75!dark!75, draw=chromeyellow!75!dark!75, fill opacity=0.5, draw opacity=0.5] coordinates {(0, 2792) (100, 2054) (500, 149) (1000, 5) (5000, 0) (30000, 0)};
    \end{axis}
\node[below, yshift=-2pt, font=\footnotesize] at (current bounding box.south) {(c) YOLO-NAS-s score threshold $=0.15$};
\end{tikzpicture}
\caption{\textbf{Distribution of the number of boxes}. The x-axis corresponds to the number of boxes that are fed into NMS. The y-axis corresponds to the number of images on COCO \texttt{val2017} whose remaining box numbers are in the corresponding interval. (a) is under the default score threshold. (b) is tuning the score threshold to get the balance between detection performance and latency. (c) is tuning a higher score threshold.}
\label{fig:boxnumdistribution}
\vspace{-3mm}
\end{figure}
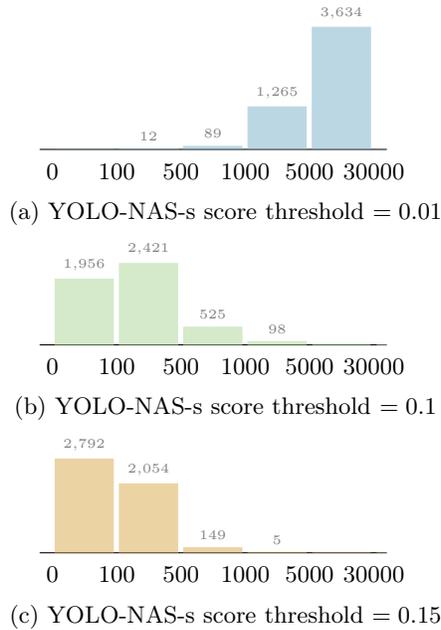

\noindent\textbf{Distribution of the number of boxes for NMS.} The latency is measured on the COCO \texttt{val2017}, which is an average of $5000$ images. Figure~\ref{fig:boxnumdistribution} gives the distribution of the number of remaining boxes in NMS across the COCO \texttt{val2017} under different score thresholds in YOLO-NAS. The large overhead brought by the NMS is due to the large number of remaining boxes under the default score threshold. Tuning the score threshold effectively decreases the remaining boxes in NMS, thus providing optimization in total latency for non-end-to-end detectors.

\end{document}